\documentclass{article}
\usepackage{spconfa4,amsmath,graphicx}
\usepackage{booktabs}
\usepackage{url}
\title{Comparison of self-supervised in-domain and supervised out-domain transfer learning for bird species recognition}
\name{Houtan Ghaffari*, Paul Devos\thanks{* Corresponding author: houtan.ghaffari@ugent.be.}}
\address{WAVES Research Group, Department of Information Technology, Ghent University, Belgium}
\begin{document}
%
\maketitle
\begin{abstract} 
Transferring the weights of a pre-trained model to assist another task has become a crucial part of modern deep learning, particularly in data-scarce scenarios. Pre-training refers to the initial step of training models outside the current task of interest, typically on another dataset. It can be done via supervised models using human-annotated datasets or self-supervised models trained on unlabeled datasets. In both cases, many pre-trained models are available to fine-tune for the task of interest. Interestingly, research has shown that pre-trained models from ImageNet can be helpful for audio tasks despite being trained on image datasets. Hence, it's unclear whether in-domain models would be advantageous compared to competent out-domain models, such as convolutional neural networks from ImageNet. Our experiments will demonstrate the usefulness of in-domain models and datasets for bird species recognition by leveraging VICReg, a recent and powerful self-supervised method.
\end{abstract}
\section{Introduction}
\label{sec:intro}
Deep learning technology is the backbone of modern data analysis in many research and application areas. One obstacle in training performant deep neural networks is the scarcity of human-annotated data for a specific task. This challenge has led to intriguing approaches for training data-efficient deep neural networks under the umbrella term of few-shot learning \cite{song2022comprehensive}. Among them, transfer learning is one well-established method \cite{zhuang2020comprehensive}, where the most ubiquitous form of this technique utilizes the pretrained parameters as a proxy to transfer the already acquired knowledge from another task or dataset. The pretrained model then requires a fine-tuning stage on a downstream task of interest, potentially with limited human-annotated data.

The result of transfer learning is often significantly superior to training randomly initialized neural networks from scratch on small to medium-sized annotated datasets. Additionally, this approach saves a great deal of computation power and time, which makes the deep learning technology accessible to a broader community. Surprisingly, transfer learning works even if the pretraining scenario is not directly related to the downstream task, such as fine-tuning the pretrained models from image classification for audio tasks \cite{DUFOURQ2022101688,gong2021ast}. However, as expected, the models pretrained on large in-domain datasets related to a downstream task would generally result in better fine-tuned models for that task \cite{ghani2023global}.

The domain of tasks is strongly related to the type of dataset used, and at a coarse level, it translates to whether we are processing text, image, audio, genomic data, etc. Furthermore, each dataset is amenable to finer in- and out-domain categorization based on its content. For example, within the audio domain, the datasets for industrial sounds, music, speech, and bird songs could be separate sub-domains that share some properties inherent to the physics of sound. A recent comprehensive study by Ghani et al. \cite{ghani2023global} compared pretrained supervised models from large-scale general audio datasets to the ones pretrained on bird sounds for few-shot species recognition. Their findings suggest that transferring in-domain supervised pretrained models were more suited for this bioacoustics task. Such results motivate research for other pretraining schemes to create general-purpose pretrained models within the bioacoustics domain, such as Self-Supervised Learning (SSL).

Unsurprisingly, the unlabeled data in any domain is vastly more than the costly human-annotated datasets. Luckily, SSL can leverage these latent sources of information without human intervention, and current SSL pretraining methods surpass the purely supervised training schemes \cite{reed2022self}. SSL relies on objective functions that do not require human-assigned targets, and they learn general domain knowledge via feature embedding. Most self-supervised methods optimize a similarity objective between differently augmented versions of data points called views since they can be seen as different views on the main content of the original data. As a result of such formulation, these methods heavily depend on the inherent property of the augmentation functions, with the hope that point-wise nullification of the dissimilarities caused by them will produce an informative representation of some fundamental semantic contents of the dataset.

Some things can go wrong with such a learning scheme, some of which are not intuitive or sufficiently comprehensible. For example, SSL objective functions might merrily encourage distancing similar samples based on human judgment/categorization. In a simplified form, this procedure is as if each datum has a unique label and is contrasted against all others except a noisy version of itself. Such formulation will disregard regularities among the semantically related samples, and it induces an optimization landscape with fundamental inconsistencies. The result is a phenomenon known as representation collapse, where flexibility and lack of inductive bias in large neural networks result in mapping all the samples into fixed or non-informative features that achieve low loss on a meaningless objective function. Hence, there are usually costly and non-intuitive heuristics at work to prevent the collapse of information, such as momentum-encoder \cite{he2020momentum,grill2020bootstrap}, stop-gradient \cite{chen2021exploring}, very large batch sizes for contrastive learning \cite{oord2019representation,pmlr-v119-chen20j}, and uniform splitting into latent categories \cite{asano2019self,NEURIPS2020_70feb62b}.

VICReg (Variance-Invariance-Covariance Regularization) \cite{bardes2022vicreg} is a recent SSL method by Bardes et al. that prevents representation collapse in a more principled way. It has a simple mean squared error term for encouraging point-wise similarity between corresponding pairs of views (invariance to augmentation). However, it does not explicitly force any specific pair of samples to be dissimilar. To prevent collapse, VICReg leverages two regularization terms that operate on individual embedding dimensions within each augmented batch of views. The first term maintains the variance of each embedding dimension within the batch above a threshold, which encourages the diversity of representation but not explicit contrasting. Therefore, although we can't control it, the model is not forced to push semantically similar samples away. The second term decorrelates each pair of different embedding dimensions within the batch to reduce the information redundancy and covariation.

We found VICReg to produce good results even with small batch sizes. Also, the hyper-parameters of the original work on image tasks were satisfactory in our bioacoustics experiments despite using different optimization settings, which is a testimony to its ease of use and stability. Therefore, we chose VICReg to investigate the utility of SSL bioacoustics models. This experimental investigation is our contribution.
\section{Data}
The SSL pretraining used the BirdCLEF2021 dataset from Kaggle challenge \cite{Kahl2021OverviewOB}, containing $\sim 63$ k recordings of various lengths from the Xeno-Canto public repository of bird sound recordings \cite{XC}. These recordings have a sampling rate of 32 ksps (kilo samples per second). Also, the ESC-50 dataset for environmental sound classification \cite{piczak2015dataset} was used for adding background noise in the data augmentation pipeline.

The downstream species classification task uses the Western Mediterranean Wetland Birds (WMWB) dataset, kindly published by Gómez-Gómez et al. \cite{GOMEZGOMEZ2023102014}. It contains 5795 annotated audio excerpts of 20 endemic bird species of the Aiguamolls de l'Empordà Natural Park and amounts to 201.6 minutes of audio. This dataset is also from Xeno-Canto, but the vocalizations of interest are strongly (precisely) annotated in their corresponding recordings. Such benchmarks are much needed in the current state of the bioacoustics field to assess the soundness of results and quality of the models vividly. We resampled WMWB recordings to 44.1 ksps and split them into 10\% train, 10\% validation, and 80\% test sets.

The experiments also use a subsampled train mini-set amounting to 10\% of the train set (1\% of total data) to assess the performance in an extreme case of label scarcity. The train and validation splits, especially the 1\% train mini-set, might be incomplete since we stratified the splits based on species labels. Although this ensures the same relative portions of classes in each set, it does not guarantee that all types of test vocalizations from each species (calls and songs) are in all sets.
\section{Methodology}
\subsection{Model architecture}
The ResNeXt-50 (32 $\times$ 4d) \cite{xie2017aggregated} model without the final classification layer is used as the backbone encoder, henceforth denoted as $f_{\theta}$, and its output has 2048 dimensions. During the SSL phase, a projection head denoted as $h_{\phi}$ is required. It simply projects the encoder output to a higher dimensional feature space, ambiguously referred to as embedding space in SSL nomenclature due to the lack of a better vocabulary to discern the outputs of $f_{\theta}$ and $h_{\phi}$. The $h_{\phi}$ consists of two sequential blocks of Linear-BatchNorm-ReLU and ends with a standalone linear layer. All three linear layers have 4096 dimensions. The $h_{\phi}$ is dropped after the pretraining phase, and we transfer the $f_{\theta}$ for finetuning on the downstream task.
\subsection{Pretraining step}\label{sec:ssl_pretrain_phase}
We denote the unlabeled pretraining dataset of size $N$ with $D_e = \{x_i\}_{i=1}^N$, where each $x_i$ is a 1-second excerpt of audio wave, randomly cropped from the full recordings of BirdCLEF2021 at each epoch $e$. An input batch of size $n$ is denoted by $X = \{x_i\}_{i=1}^n$. The $\mathcal{T}$ denotes the data augmentation and transformation pipeline, consisting of the following steps in order:
\begin{itemize}
    \item \textbf{pitch shifting}: shift steps on the scale of 12 steps per octave were chosen randomly from the discrete range of [-4, 4].
    \item \textbf{background noise}: adds a random recording from the ESC-50 dataset \cite{piczak2015dataset} as background noise with a signal-to-noise ratio (SNR) randomly chosen from the discrete range of [1, 20].
    \item \textbf{Short-Time Fourier-Transform (STFT)}: transforms waveforms to time-frequency representation with an fft size of 800 points (25 ms at 32 ksps) and a hop size of 320 (10 ms).
    \item \textbf{time-stretching} \cite{park2019specaugment}: stretches/shrinks the STFT in time without modifying pitch for a stretch rate randomly chosen from the real interval of [0.9, 1.1].
    \item \textbf{log compression}: transforms the STFT values from magnitude to decibel scale.
    \item \textbf{min-max normalization}: normalizes each data point (not batch) to the real range [0, 1].
    \item \textbf{time and frequency masking} \cite{park2019specaugment}: a random time mask of a maximum of 8 frames and a frequency mask of a maximum of 16 bins hide portions of the inputs.
\end{itemize}

We apply $\mathcal{T}$ twice on a batch of waveform input $X$ to get two augmented batches of spectrogram views $X'$ and $X''$ in $\mathcal{R}^{n \times F \times T}$. The encoder maps the two views into $Y' = f_{\theta}(X')$ and $Y'' = f_{\theta}(X'')$ in $\mathcal{R}^{n \times d'}$, and the projection head takes them to embedding space $Z' = h_{\phi}(Y')$ and $Z'' = h_{\phi}(Y'')$ in $\mathcal{R}^{n \times d}$. The VICReg objective function consists of three terms that operate on these two embeddings \cite{bardes2022vicreg}.

The first loss term is the invariance (to augmentation) criterion between the corresponding pairs of embeddings from the two views. It encourages the model to map semantically similar samples closer to each other: 
\begin{equation}
    s(Z', Z'') = \frac{1}{n} \sum_{i=1}^{n} ||z'_i - z''_i||_2^2
\end{equation}

The second term is variance regularization. It is calculated in each batch of views separately to prevent degenerate constant embeddings. Using the standard deviation is necessary to avoid embedding collapse since the gradient of variance function for inputs close to the center is near zero \cite{bardes2022vicreg}: 
\begin{equation}
    v(Z) = \frac{1}{d} \sum_{j=1}^{d} max(0, \gamma - \sqrt{Var(z^j) + \epsilon})
\end{equation}
where $Z \in \mathcal{R}^{n \times d}$ is either batch of views ($Z'$ and $Z''$), $z^j \in \mathcal{R}^n$ and superscript $j$ indexes the feature dimension, $\epsilon$ is a small constant to avoid numerical instabilities, and $\gamma$ is a constant target value for the standard deviation. Following the original work \cite{bardes2022vicreg}, we set $\epsilon=0.0001$ and $\gamma=1$.

The third term is covariance regularization. It pushes the covariances to zero to reduce redundancy in embedding dimensions by decorrelating them and preventing information collapse \cite{bardes2022vicreg}. The covariance matrix of embeddings is:
\begin{align}
    C(Z) & = \frac{1}{n-1} \sum_{i=1}^{n} (z_i - \Bar{z})(z_i - \Bar{z})^T\\
    \Bar{z} & = \frac{1}{n} \sum_{i=1}^{n} z_i
\end{align}
and the loss for regularizing it is defined as:
\begin{equation}
    c(Z) = \frac{1}{d} \sum_{i \neq j} |C(Z)|^2_{i,j}
\end{equation}

The overall loss function is a weighted average of these three terms:
\begin{equation}
    \ell(Z', Z'') = \lambda s(Z', Z'') + \mu [v(Z') + v(Z'')] + \nu [c(Z') + c(Z'')]
\end{equation}
we used $\lambda = 25$, $\mu=25$, and $\nu=1$, same as the \cite{bardes2022vicreg}.

The model was trained for 100 epochs using the SGD optimizer with a momentum of 0.9 and weight decay of $1\mathrm{e}{-4}$. A learning rate scheduler linearly increased it from $1\mathrm{e}{-4}$ to $0.3$ in 10 epochs and then reduced it with a cosine decay curve to a minimum of $1\mathrm{e}{-4}$. The batch size was 192.
\subsection{Species classification}
The downstream task is to classify the 20 species from the WMWB dataset \cite{GOMEZGOMEZ2023102014} using limited labels. Three models were compared in two scenarios, and all three are based on the ResNeXt-50 (32 $\times$ 4d) \cite{xie2017aggregated} model after changing the classification layer to a randomly initialized linear layer with 20 outputs. One model was randomly initialized and trained from scratch. The second model initialized its encoder part with the pretrained weights from the ImageNet classification task \cite{imagenet_cvpr09}. The third model used our SSL pretrained encoder as described in section \ref{sec:ssl_pretrain_phase}. All models were trained in two cases of 10\% and 1\% labeled data. The two pretrained models were also compared in linear probe condition \cite{alain2018understanding}. It means freezing the encoder and only training a classifier on top of the fixed features to assess their adequacy for the downstream task.

All models were trained for 100 epochs using the Adam optimizer. A learning rate scheduler linearly increased it from $1\mathrm{e}{-5}$ to $1\mathrm{e}{-3}$ in 10 epochs and then reduced it with a cosine decay curve to a minimum of $1\mathrm{e}{-5}$. The data augmentation was not used here and the waveforms were transformed into spectrograms using the same STFT setup as section \ref{sec:ssl_pretrain_phase}, log compression, and min-max normalization. The experiments were run five times to report the mean and standard deviation of the results.
\section{Results}
Table \ref{tab:test_1} shows the test results after using the train mini-set (1\% of the dataset). Table \ref{tab:test_10} is the same using the full training set. The test and validation sets were the same in both cases. ImageNet pretrained models are strong baselines in transfer learning, and it is challenging to outperform them significantly. However, it is evident that SSL pretraining using in-domain bioacoustics data far surpasses the ImageNet pretrained model. Also, notice: 1) ImageNet is more than 20 times bigger than the dataset we used ($\sim 1.3$ M), and 2) its pretrained models are from supervised learning using human annotation.

\begin{table}[!t]
\setlength\tabcolsep{3pt}
 \caption{Test results using train mini-set (1\% of the WMWB dataset). The code 'L' means linear probe, where the encoder is frozen and only the linear classifier gets trained. The code 'F' means finetuned where the whole model is trained and optimized on the train set. There is no pretrained encoder in the random initialization case, thus, we did not do a linear probe. The mean and standard deviation are from five different runs.}
 \resizebox{\columnwidth}{!}{\begin{tabular}{l*{6}{c}}
    \toprule
    Model     & accuracy & top-3 accuracy & f1-score & precision & recall\\
    \midrule
    rand-init & $54.3 \pm 1.3$ & $77.1 \pm 1.1$ & $45.5 \pm 1.4$ & $48.0 \pm 1.7$ & $44.9 \pm 1.4$ \\
    imgnet-L  & $33.8 \pm 0.1$ & $66.7 \pm 0.2$ & $23.1 \pm 0.3$ & $43.0 \pm 0.4$ & $20.5 \pm 0.1$ \\
    vicreg-L  & $52.3 \pm 0.3$ & $72.5 \pm 0.3$ & $34.2 \pm 0.2$ & $45.0 \pm 1.7$ & $32.1 \pm 0.2$ \\
    imgnet-F  & $59.5 \pm 1.4$ & $82.6 \pm 0.6$ & $50.8 \pm 1.5$ & $57.8 \pm 1.3$ & $49.8 \pm 1.8$ \\
    vicreg-F  & $77.3 \pm 1.0$ & $91.0 \pm 0.5$ & $70.7 \pm 0.9$ & $72.0 \pm 1.1$ & $70.4 \pm 0.9$ \\
    \bottomrule
 \end{tabular}}
 \label{tab:test_1}
\end{table}

\begin{table}[!t]
\setlength\tabcolsep{3pt}
 \caption{Same as Table \ref{tab:test_1} but using the full train set (10\% of the WMWB dataset).}
 \resizebox{\columnwidth}{!}{\begin{tabular}{l*{6}{c}}
    \toprule
    Model     & accuracy & top-3 accuracy & f1-score & precision & recall\\
    \midrule
    rand-init & $90.7 \pm 0.3$ & $97.9 \pm 0.1$ & $87.9 \pm 0.6$ & $88.2 \pm 0.6$ & $87.8 \pm 0.6$ \\
    imgnet-L  & $61.4 \pm 0.2$ & $85.0 \pm 0.2$ & $54.8 \pm 0.2$ & $60.7 \pm 0.3$ & $51.9 \pm 0.4$ \\
    vicreg-L  & $76.3 \pm 0.3$ & $94.4 \pm 0.1$ & $70.5 \pm 0.3$ & $72.3 \pm 0.2$ & $69.7 \pm 0.3$ \\
    imgnet-F  & $93.5 \pm 0.4$ & $98.6 \pm 0.1$ & $91.9 \pm 0.4$ & $92.1 \pm 0.5$ & $91.8 \pm 0.3$ \\
    vicreg-F  & $94.3 \pm 0.4$ & $99.0 \pm 0.1$ & $92.7 \pm 0.6$ & $92.6 \pm 0.7$ & $92.8 \pm 0.5$ \\
    \bottomrule
 \end{tabular}}
 \label{tab:test_10}
\end{table}
\section{Conclusion}
\label{sec:conclusion}
This work showed the advantage of in-domain transfer learning via self-supervised pretraining compared to the main-stream out-domain ImageNet supervised pretrained weights. A successful application of SSL often requires a large amount of data to produce adequate results. However, the experiments showed that an in-domain SSL model for bird species recognition is invaluable, even with much less data than today's standards. Hence, this is a promising result to motivate future research in building larger SSL models for bioacoustics. It will benefit the community for many bioacoustics tasks that require feature extraction from audio recordings.

Reliable and competent pretrained models can stabilize and unify parts of future works and make the results easier to compare and reproduce. Also, adequate pretrained models will save computational resources and make a performant data analysis pipeline accessible to a broader audience. Furthermore, this procedure will reduce development time since finetuning a robust model is much less challenging than building one from scratch. The latter requires numerous trials and errors to create a satisfying model.

\bibliographystyle{IEEEbib}
\bibliography{refs}

\end{document}